\documentclass[11pt]{article}

\usepackage[preprint]{acl}

\usepackage{times}
\usepackage{latexsym}
\usepackage{arydshln}
\usepackage{booktabs}

\usepackage[T1]{fontenc}

\usepackage[utf8]{inputenc}

\usepackage{microtype}

\usepackage{inconsolata}

\usepackage{graphicx}
\usepackage{subfig}
\usepackage{amsmath, xcolor}
\usepackage[most]{tcolorbox}
\usepackage{listings}
\usepackage{xcolor}
\lstdefinestyle{prompt}{
  basicstyle=\small\ttfamily,
  breaklines=true,
  breakatwhitespace=false,
  columns=fullflexible,
  keepspaces=true,
  frame=none,
}

\newcommand{\blue}[1]{\textcolor{black}{#1}}

\title{TACO: Task-Aware Column Description Generation Using Large Language Models}

\author{
  \textbf{Ting Cai\textsuperscript{1}},
  \textbf{Rakesh R. Menon\textsuperscript{2}},
  \textbf{Yiru Chen\textsuperscript{2}},
  \textbf{Zifan Liu\textsuperscript{2}},
\\
  \textbf{Yuan Tian\textsuperscript{3}},
  \textbf{Fei Wu\textsuperscript{2}},
  \textbf{Anudeep Chimakurthi\textsuperscript{2}},
  \textbf{Prashanthi Ramamurthy\textsuperscript{2}},
\\
  \textbf{Sunav Choudhary\textsuperscript{2}},
  \textbf{Kun Qian\textsuperscript{2}},
  \textbf{Yunyao Li\textsuperscript{2}}
\\
\\
  \textsuperscript{1}University of Wisconsin-Madison
  \textsuperscript{3}Purdue University,
  \textsuperscript{2}Adobe,
\\[6pt]
  \begin{minipage}{0.9\linewidth}
    \centering
    \small
    \texttt{tian211@purdue.edu}, \texttt{tingcai@cs.wisc.edu}, \texttt{\{rakeshra, yiruc, zifanl, feiw, achimakurthi, pramamur, schoudha, kunq, yunyaol\}@adobe.com}
  \end{minipage}
}

\begin{document}

\maketitle
\begin{abstract}

Structured data is ubiquitous, and large language models (LLMs) are increasingly leveraged to reason over such data in tasks such as entity linking, table question answering, and natural language-to-SQL translation. 
However, opaque and uninformative column names—prevalent in real-world datasets—limit the ability of LLMs to identify correct columns, thereby hindering semantic reasoning. To support effective semantic retrieval and understanding, columns must be enriched with accurate, context-aware descriptions. We present TACO, an automatic three-stage framework for task-aware column description generation. TACO generates enriched descriptions incorporating domain-relevant cues and refines them by optimizing their utility for downstream reasoning tasks.
Extensive experiments across public and proprietary datasets show that TACO consistently outperforms existing methods, improving downstream entity linking performance by up to 52\%. 
We further demonstrate that incorporating user feedback with TACO yields additional performance gains.

\end{abstract}

\section{Introduction}
\begin{figure}[t]
    \centering
    \includegraphics[width=1\linewidth]{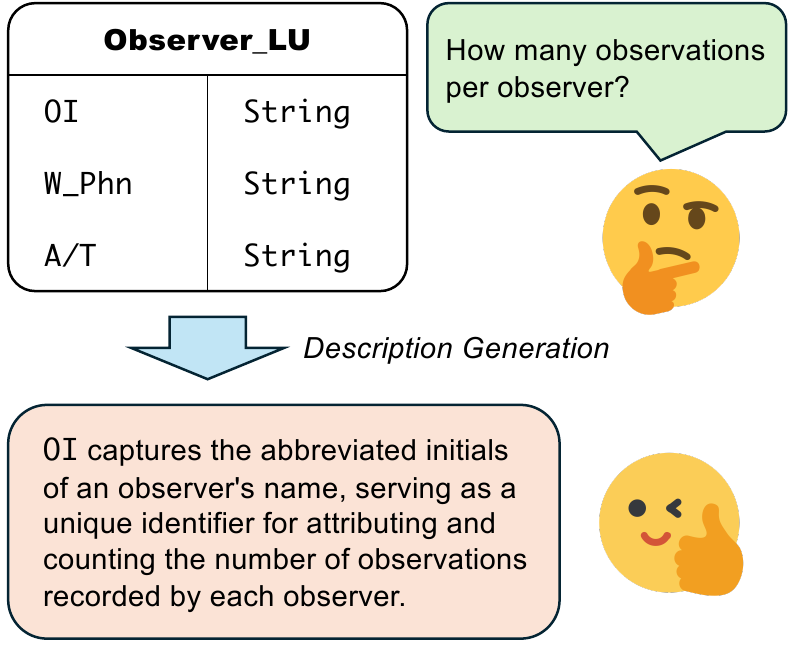}
    \caption{\blue{An illustrative example of column description generation, where the generated description provides enough context to choose the correct column for answering the question.}}
    \label{fig:intro}
\end{figure}
Structured data is ubiquitous across enterprise, governmental, and scientific domains.
A wide range of tasks, such as entity linking, table question answering, keyword search, and natural language–to–SQL (NL2SQL) translation, require accurate interpretation of the schema semantics.
However, in practice, the vast majority of datasets provide only minimal or cryptic metadata.
Column names are often abbreviated (e.g., \texttt{esal} for ``employee salary") or domain-specific,
while column descriptions are often missing altogether.
For instance, our analysis of five large enterprise datasets (over tens of thousands of columns) from a data vendor,
revealed that over 95\% of columns lacked descriptions,
and more than 40\% contained abbreviations that were difficult to interpret.
Even widely used human-curated benchmarks, such as BIRD~\cite{birdbench, wretblad2024synthetic},
exhibit this issue; e.g., there are tables containing opaque column names such as \texttt{A2} without any accompanying descriptions.

Semantic enrichment, particularly through column description generation, is essential for improving the interpretability of relational data and its usability in conversational systems.
Figure~\ref{fig:intro} shows an example of the column description generation task.
For example, given the column \texttt{OI} in a scientific observation table, a generated description might state ``\texttt{OI} captures the abbreviated initials of an observer's name, serving as a
unique identifier  ...''.
Such metadata can substantially aid downstream tasks: an NL2SQL
system can answer the query ``How many observations
per observer?'' far more effectively when provided with the description rather than just the column name \texttt{OI}.

Recent advances in large language models (LLMs) have inspired efforts to automatically enrich table semantics by generating column descriptions.
A common approach
is to prompt LLMs with table and column names (and optionally data values) to produce column descriptions~\cite{wretblad2024synthetic,zhang2025autoddg}.
However, existing LLM-based solutions, which rely on few-shot prompting and operate in a single pass, exhibit three fundamental limitations.
First, LLMs often interpret abbreviations inconsistently, expanding the same abbreviation differently across columns within the same dataset.
For example, the abbreviation ``wt'' may be interpreted as ``witness'' in \texttt{wt\_species\_code} but as ``weight'' in \texttt{wt\_stake}, even though both should share the same meaning within the same table.
Second, generated descriptions frequently introduce hallucinations or irrelevant details.
Third, the descriptions are often noisy or overly generic, with key semantic cues missing.
Together, these issues limit the effectiveness of the descriptions in downstream tasks.

To overcome these challenges, we propose TACO (\textbf{T}ask-\textbf{A}ware \textbf{Co}lumn description generation), a novel
three-step framework:

\noindent \textbf{Column name expansion} resolves abbreviations prior to description generation, ensuring consistent interpretation across columns. Expanding them jointly over the full dataset also provides richer semantic signals.

\noindent \textbf{Description generation} produces semantically rich descriptions guided by table context. In addition, the descriptions are augmented with synonyms, keywords, and proxy search queries for better semantic retrieval.

\noindent \textbf{Description revision} simulates downstream tasks (e.g., semantic retrieval) to refine and disambiguate descriptions based on their performance in these tasks, generating task-aware descriptions that remedy overly generic LLM outputs and missing semantic cues.

As an extension, TACO can elicit human feedback for abbreviation expansions in which it exhibits low confidence. The resulting human-provided corrections are then propagated across the dataset to refine column descriptions, enabling TACO to flexibly adapt to a human-in-the-loop setting.

Our contributions are as follows:
\begin{itemize}
    \item We introduce TACO, a comprehensive three-step solution that produces consistent, semantically rich, and task-aware column descriptions through dataset-wide column name expansion, context-guided description generation, and downstream-task-informed revision.
    \item Through extensive experiments on both public benchmarks and proprietary datasets for the task of entity linking, we demonstrate that TACO outperforms state-of-the-art baselines, achieving up to 52\% improvements in retrieval performance.
    \item We provide a new evaluation dataset to facilitate and benchmark research on task-aware column description generation.
\end{itemize}

\section{Problem Statement}
We formally define the task of column description generation as follows.
Let a dataset be represented as a collection of tables $\mathcal{T} = \{t_1, t_2, \cdots, t_m\}$, 
where $m$ is the number of tables in the dataset.
Each table
$t_i = (n_i, C_i)$ consists of a table name $n_i$ and 
a set of column names $C_i = \{c_{i1}, c_{i2}, \cdots, c_{iu}\}$, where $u$ is the number of columns in the table.

The objective is to generate, for every column $c_{ij}$, 
a natural language description $d_{ij}$ that conveys its semantics and facilitates downstream tasks
such as NL2SQL, keyword search, table QA and entity linking.
Formally, we seek a mapping
\begin{equation}
    f_{\theta}: (n_i, C_i, \mathcal{T}) \rightarrow \{d_{i1}, d_{i2}, \cdots, d_{iu}\},
\end{equation}
where $\theta$ denotes the parameters of the model used to generate the descriptions.
Unlike prior work that directly applies a single prompt to compute $d_{ij}$, 
we decompose $f_{\theta}$ into three stages:
(1) \textbf{Abbreviation expansion} produces expanded forms $\hat{c}_{ij}$ for each column name $c_{ij}$ and $\hat{n}_i$ for each table name $n_i$.
(2) \textbf{Description generation} generates candidate descriptions $\tilde{d}_{ij}$ conditioned on expanded column and table names.
(3) \textbf{Description revision} refines $\tilde{d}_{ij}$ into final description $d_{ij}$ by incorporating feedback from simulated downstream tasks.


\blue{We restrict our inputs to metadata, namely table and column names, and assume no access to column values. This choice is motivated by data privacy and compliance constraints, as raw database values may contain sensitive information subject to regulation~\cite{privacy_regulation2016regulation, privacy_illman2019california} and are often prohibited from being exposed to learning systems. As a result, metadata-only access is a common assumption in prior schema linking and NL2SQL works~\cite{schema1_zhang2023schema, tian-etal-2023-interactive, schema2_lei2020re, 10.1145/3654777.3676368, schema3_maamari2024death}.}

\section{Method}

\begin{figure}[t]
    \centering
    \includegraphics[width=1\linewidth]{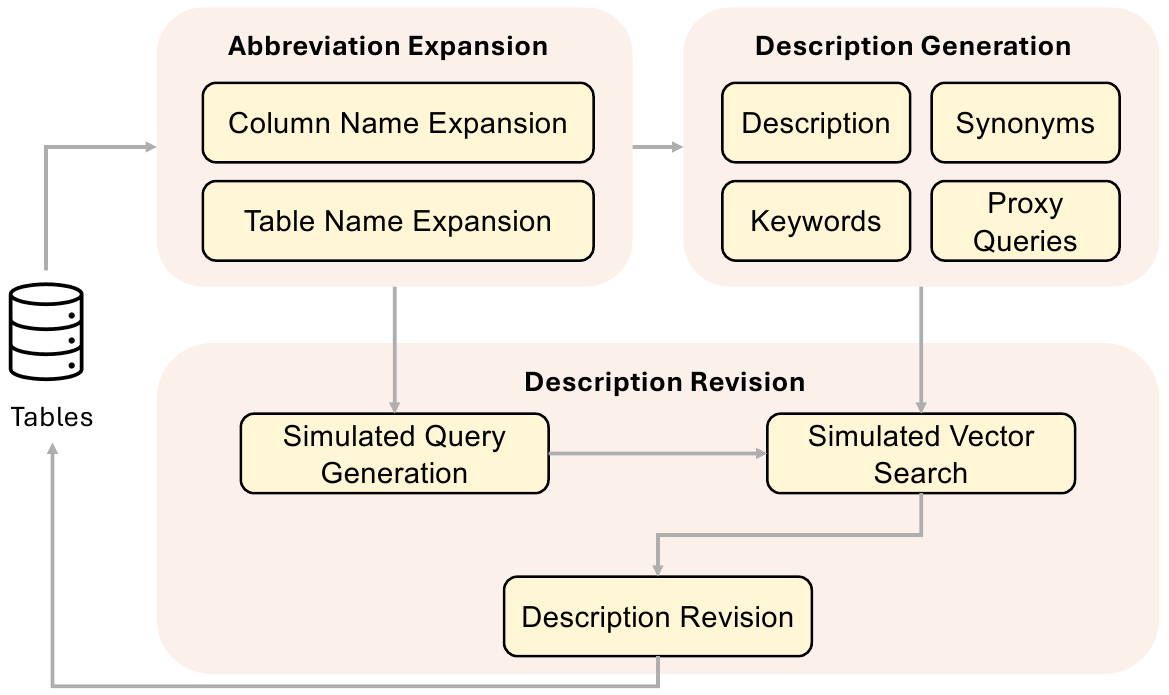}
    \caption{\blue{Overview of TACO. TACO has three main stages: Abbreviation Expansion, Description Generation, and Description Revision.}}
    \label{fig:solution}
\end{figure}

In this section, we present \textbf{TACO}, a modular framework for task-aware column description generation that addresses three limitations of single-prompt baselines: inconsistent abbreviation interpretation, hallucinated or weakly grounded descriptions, and a lack of downstream awareness. As illustrated in Figure~\ref{fig:solution}, TACO consists of three sequential stages. \emph{Abbreviation expansion} performs dataset-level expansion to ensure consistent interpretation across columns. Next, \emph{Description generation} conditions on enriched schema context and expanded column names, reducing reliance on opaque identifiers and mitigating hallucinations. Finally, \emph{Description revision} leverages semantic retrieval as a downstream proxy to refine descriptions based on task performance, explicitly injecting downstream awareness.

\subsection{Abbreviation Expansion}
In this section, we explain how we expand the abbreviated column names and table names into schema-consistent, semantically grounded name expansions.

\paragraph{Column name expansion.}

We expand each abbreviated column name $c_{ij}$ into a schema-consistent natural language expansion $\hat{c}_{ij}$ using a LLM-based component that conditions on schema context, including other columns within the same table and related tables. 
We implement this component using
the \textsc{Columbo} method~\cite{columbo}, which (i) exploits intra-table and inter-table column context, (ii) performs token-level chain-of-thought expansion, and (iii) 
enforces dataset-level consistency so that the same abbreviation is expanded uniformly across the schema.

\paragraph{Table name expansion.}
We also expand each table name $n_i$ into a concise natural language title $\hat{n}_i$ using the expanded columns as context.
When a table contains more than $k$ columns, we randomly sample $k$ columns to control prompt length. 
The full prompt to generate the table expansion can be found in Appendix~\ref{sec:table name expansion prompt}.
We enforce ground expansions by conditioning the prompt on 
information implied by the columns and prohibiting the introduction of new domain terms or external explanations.

\subsection{Description Generation}

Given the expanded table name and expanded column names, we generate column descriptions using a prompt applied to groups of $K$ columns per table. Each prompt conditions on the expanded table name together with the expanded names of the $K$ columns. To ensure grounded and consistent descriptions, we include explicit constraints in the prompt; the full prompt is provided in Appendix~\ref{sec:column description prompt}. Specifically, the prompt enforces the following rules: (i) disambiguate columns with similar names within the same table, (ii) avoid inventing fields or data, and (iii) prefer domain-agnostic phrasing unless the table context clearly indicates otherwise.

Each generated description is structured to support semantic retrieval and consists of four components: (i) a one-sentence overview, (ii) a set of aliases, (iii) a set of relevant keywords, and (iv) a collection of proxy user queries that approximate potential search intents for retrieving the column. As an example, the description generated for \texttt{OI} includes the following:\vspace{.3em}
\fbox{\parbox{0.99\linewidth}{
\noindent \textbf{Overview:} ``\texttt{OI} captures the abbreviated initials of an observer's name ...''

\noindent \textbf{Aliases:} ``observer code'', ``scientist initials''

\noindent \textbf{Keywords:} ``unique ID'', ``observer initials''

\noindent \textbf{Proxy Queries:}  ``identify observer by initials'', ``find observer abbreviation in data''
}}

\subsection{Description Revision}
\label{sec:revision}
\begin{figure}[h]
    \centering
    \includegraphics[width=1\linewidth]{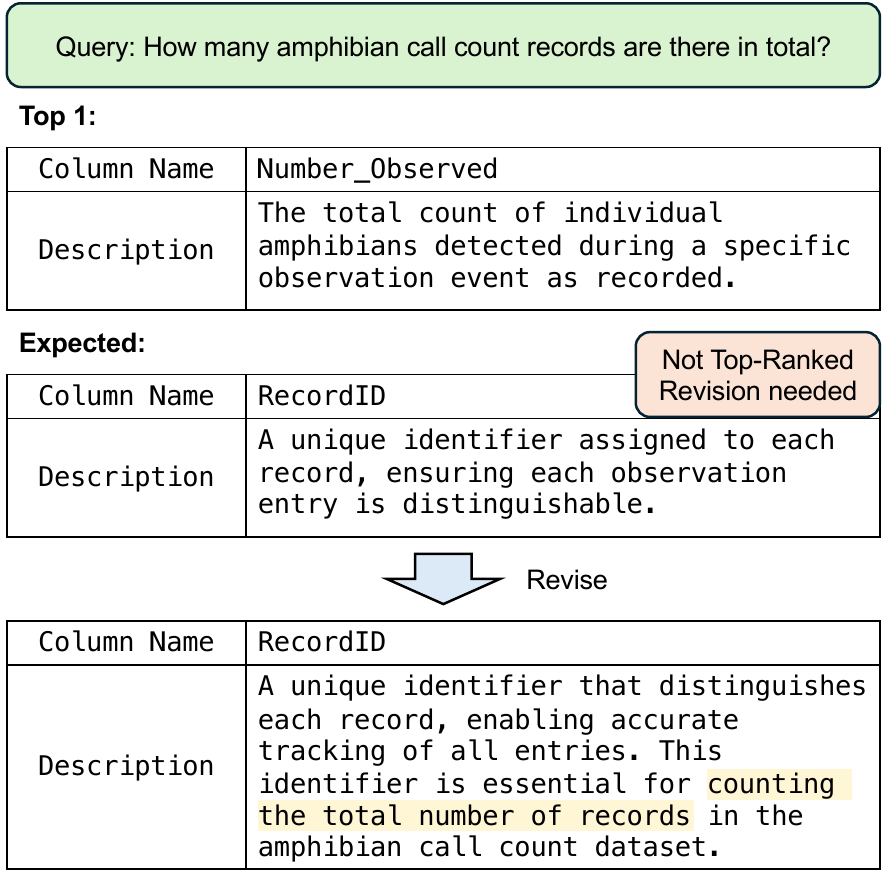}
    \caption{\blue{An example of description revision. Aliases, keywords, and proxy user queries are revised as well, but are not shown for conciseness.}}
    \label{fig:example}
\end{figure}
After abbreviation expansion and structured description generation, some columns remain difficult to disambiguate due to overlapping semantics.
In particular, columns may share similar surface forms or refer to closely related concepts. Figure~\ref{fig:example} illustrates such a case, where two columns refer to closely related but different concepts, leading to an incorrect top-1 retrieval.
To address this issue, 
we design a revision stage that explicitly incorporates downstream retrieval behavior. The key idea is to simulate semantic retrieval and 
use the retrieval failures to guide refinement.
Specifically, for each column we prompt the LLM to generate plausible user queries,
perform vector search over all candidate columns using the generated descriptions,
and detect cases where the intended column (the column used to generate the query) 
is not ranked first. 
In such cases, we gather the descriptions of the top competing columns together with the intended column and prompt the LLM to revise the intended column to make the description more distinctive and better aligned with the query. 
This process encourages the revised descriptions to 
reduce overlap among semantically similar columns and ultimately 
improves retrieval performance in downstream tasks, as shown in the experiment section (Section 6).

More precisely, for each column $c$ with description $\tilde{d}_c$, 
we instruct the LLM to first generates $M$ (default to be $3$) plausible user queries $\{q^{(1)},\cdots,q^{(M)}\}$. 
Each query $q^{(m)}$ is embedded and compared via cosine similarity against the embeddings of all candidate column descriptions, including the one-sentence overview, aliases, keywords, and proxy queries.
We consider the case as a retrieval error if the highest-scoring description does not correspond to the intended column 
$c$, i.e.,
\begin{equation}
    \arg\max_{c'} r(q^{(m)}, \tilde{d}_{c'}) \neq c,
\end{equation}
where $r(q,d)=\langle \phi(q),\phi(d) \rangle$ is the cosine similarity of the embeddings,
and $\phi$ is the embedding function.
We collect the top $V$ competing columns for that query and ask the LLM to revise $\tilde{d}_c$
into a more distinctive description $\tilde{d'}_c$.
For efficiency, we group all queries associated with the same column into a single prompt and ask the LLM to perform a single revision per column.

Finally, we evaluate the effectiveness of the revision by comparing the average cosine similarity of the original and revised descriptions against all simulated queries:
\begin{equation*}
    \frac{1}{M}\sum_{m=1}^M r(q^{(m)}, \tilde{d}_c) \quad \textit{vs.} \quad \frac{1}{M}\sum_{m=1}^M r(q^{(m)}, \tilde{d'}_c).
\end{equation*}
The final description $d_c$ is set to $\tilde{d'}_c$ only if it achieves a higher average similarity
than the original $\tilde{d}_c$.
Otherwise,
we retain the existing description.
Prompts for both query generation and description revision are provided in Appendix~\ref{apx:column revision}.

\section{Human-in-the-Loop Extension}

This section considers how additional human input can be incorporated
to improve name expansions.
This serves as an initial demonstration, and a more comprehensive study of human-in-the-loop strategies 
is left to future work. The prompts mentioned in this section can be found in Appendix~\ref{apx:sec:prompt human input}.

\subsection{Identification of Expansions Requiring Human Verification}

To automatically identify low-confidence expansions for human verification, we leverage an LLM as a judge.
Given a table, its expanded table name, a column name, and a candidate column name expansion, we prompt the LLM to assess whether the expansion is likely to be correct in the table context and outputs a score on a 0 to 5 scale, where 0 indicates very unlikely and 5 indicates very likely to be correct. These scores are used to automatically flag low-confidence cases that need human verification.

\subsection{Human-Revised Expansion Propagation}

When human users provide verified expansions for specific table and column names, this feedback can be propagated to improve the quality of other expansions. To integrate such feedback at scale while avoiding LLM context-window limitations, we adopt a retrieval-based strategy. Specifically, for a target column, we retrieve two types of curated feedback: (i) verified expansions from other columns within the same table, and (ii) verified expansions from columns with the same name in other tables. These retrieved examples are then incorporated into the prompt to guide the LLM in revising candidate expansions.


\section{Evaluation Dataset}

\begin{table}[t]
    \centering
    \small
    \begin{tabular}{c|c|c|c|c}
        \toprule
       source & dataset   & \# tables & \# columns & \# queries  \\
       \midrule
        & ASIS  & 36 & 245 & 245 \\
        & ATBI & 28 & 192 & 192 \\
        & CWO & 13 & 71 & 71\\
        & KIS & 18 & 157 & 157 \\
       SNAILS & NPFM & 27 & 190 & 190 \\
        & NTSB & 40 & 1,611 & 1,611 \\
        & NYSED & 27 & 423 & 423\\
        & PILB & 21 & 196 & 196 \\
        & SBOD$^*$ & 2,588 & 90,477 & 90,477 \\
       \midrule
        & Retail & 669 & 5,241 & 2,102 \\
       Enterprise & Finance & 88 & 1,291 & 735 \\
        & Tech & 127 & 10,227 & 6,470 \\
       & Tech-H & 127 & 10,227 & 78 \\
       \bottomrule
    \end{tabular}
    \caption{Statistics of the evaluation datasets. SBOD includes nine subdatasets (see Appendix~\ref{tab:sbod dataset}). Tech and Tech-H share the same table inputs, while Tech-H contains more challenging queries.}
    \label{tab:dataset}
\end{table}

We describe the evaluation sets for our experiments.
Each dataset contains the table names, associated column names, and a set of gold query-column pairs.
Dataset statistics are in Table~\ref{tab:dataset}.
We focus on retrieving target columns referred to by natural language queries, which is a foundational step in NL2SQL, schema linking, and data discovery.

\paragraph{Public Dataset}
We construct public evaluation datasets based on SNAILS~\cite{snails}, an NL2SQL benchmark that includes column names with varying levels of abbreviation and corresponding ground-truth name expansions.
We create gold query-column pairs for keyword search as follows.
For each table, we first use the gold expansions of the table and column names 
and prompt the LLM to generate a potential user query targeting that column. 
For example, given the table name \texttt{emp\_info} and column name \texttt{e\_sal},
the generated query would be ``what is the average salary for the employee in the company?''
We provide in-context examples drawn from the SNAILS gold natural language and SQL pairs as a reference to generate the queries. 
We then manually inspect and verify the correctness and soundness of the generated column-query pair.

\paragraph{Proprietary Dataset}
We use four enterprise datasets obtained from a data vendor, namely Retail, Finance, Tech, Tech-H that comes from different domains, where Tech and Tech-H use the same table schemas,
while Tech-H has a harder set of queries.
The table schemas come from the real customer data, and the gold query-column sets are generated by the data vendors.

We evaluate on four proprietary enterprise datasets provided by a data vendor, namely Retail, Finance, Tech, and Tech-H, which span multiple domains. Tech and Tech-H share the same table schemas, while Tech-H contains a more challenging set of queries. The table schemas are derived from real customer data, and the gold query column annotations are generated by the data vendor.

\section{Experiments}

We conduct extensive experiments across public and proprietary enterprise datasets to evaluate the effectiveness of TACO on semantic column retrieval. In addition, we conduct sensitivity analyses and ablation studies, and evaluate the effectiveness of the human-in-the-loop extension.

\paragraph{Experimental Setup and Metrics}
We use vector search as the primary downstream task to evaluate retrieval performance. Specifically, natural language queries are embedded and matched against vector representations formed by concatenating \texttt{table\_name.column\_name} with the column description, and performance is evaluated by the system’s ability to retrieve the intended columns. \blue{This setting reflects real-world scenarios where vector-based semantic retrieval serves as the core mechanism for schema grounding and data discovery. We use column retrieval performance to evaluate the contribution of description generation independently of other components in end-to-end tasks such as NL2SQL.}
We report
\textbf{Hit@K (H@K)}, which shows whether the correct column appears in the top $K$ results.

\paragraph{Methods} 

The most closely related work is by \citet{wretblad2024synthetic}, which uses a single prompt to directly generate column descriptions with an LLM. Several other studies on metadata generation \cite{zhang2025autoddg, gao2025automaticdatabasedescriptiongeneration, singh2025leveragingretrievalaugmentedgenerative, tabmeta} also rely on LLM-generated column descriptions, but primarily treat them as an intermediate step for table-level description generation rather than as a standalone retrieval artifact.
These approaches largely follow a prompt-only paradigm similar to \citet{wretblad2024synthetic}. Accordingly, we use the method of \citet{wretblad2024synthetic} as the representative baseline for prompt-only column description generation. In addition, we introduce our own baseline, S2-only, which applies the prompt from Step 2 of our pipeline to directly generate column descriptions, without abbreviation expansion or description revision. In summary, the evaluated methods are as follows:
\noindent \textbf{SSCD}~\cite{wretblad2024synthetic}: A LLM-based approach that generates column descriptions using a single prompt.

\noindent \textbf{S2-only}: A simplified variant of TACO that applies Step 2 (description generation) only, without abbreviation expansion or description revision.

\noindent \textbf{TACO}: The full TACO pipeline, including name expansion, description generation, and description revision. \blue{Note that no human-in-the-loop input is used unless otherwise stated in Section~\ref{label:human}.}

We use GPT-4o for LLM-based generation with temperature set to 0 and all other parameters at their default values, and all-MiniLM-L12-v2~\cite{reimers-2019-sentence-bert} for embedding generation.

\subsection{Retreival Performance}
We compare TACO against SSCD and S2-only on the evaluation datasets constructed from SNAILS and proprietary enterprise data. We also include Raw Schema as a reference point, which performs retrieval by embedding only the concatenation of the table and column names (e.g., \texttt{table\_name.column\_name}).
From Table~\ref{tab:main concat}, we can see that all three methods outputperform Raw Schema, which shows that LLM-generated descriptions are helpful in the downstream retrieval task.
Moreover, TACO consistently outperforms SSCD across all experiments and metrics, by 3-38\% absolute accuracy on $H@1$, 0-49\% on $H@5$, and 0-52\% on $H@10$, demonstrating more effective description generation. We further find that TACO performs substantially better than or comparable to S2-only, underscoring the benefits of column name expansion and description revision.

\begin{table*}[t]
    \centering
    \small
    \begin{tabular}{c|c|c|c|c|c|c|c|c|c|c|c|c}
    \toprule
     dataset    &  \multicolumn{3}{c|}{Raw schema} & \multicolumn{3}{c|}{SSCD} & \multicolumn{3}{c|}{S2-only} & \multicolumn{3}{c}{TACO (ours)} \\
     & H@1 & H@5 & H@10 
     & H@1 & H@5 & H@10 
     & H@1 & H@5 & H@10 
     & H@1 & H@5 & H@10 \\
     \midrule
     ASIS 
     & 0.05 & 0.12 & 0.16 
     & 0.13 & 0.22 & 0.26 
     & 0.26 & 0.47 & 0.58 
     & \textbf{0.29} & \textbf{0.54} & \textbf{0.68} \\ 
     ATBI 
     & 0.11 & 0.20 & 0.29 
     & 0.10 & 0.24 & 0.34 
     & 0.33 & 0.56 & 0.66 
     & \textbf{0.43} & \textbf{0.71} & \textbf{0.82}
     \\
     CWO 
     & 0.07 & 0.32 & 0.45 
     & 0.06 & 0.28 & 0.38 
     & \textbf{0.31} & \textbf{0.72} & \textbf{0.85}
     & 0.30 & 0.69 & 0.83 \\
     KIS
     & 0.07 & 0.21 & 0.29
     & 0.15 & 0.29 & 0.35
     & 0.38 & 0.64 & 0.78
     & \textbf{0.53} & \textbf{0.78} & \textbf{0.87} \\
     NPFM 
     & 0.08 & 0.22 & 0.29 
     & 0.11 & 0.21 & 0.27
     & 0.25 & 0.46 & 0.54
     & \textbf{0.42} & \textbf{0.66} & \textbf{0.74} \\
     NTSB 
     & 0.03 & 0.09 & 0.12 
     & 0.06 & 0.12 & 0.14
     & \textbf{0.20} & 0.38 & 0.47
     & 0.18 & \textbf{0.39} & \textbf{0.48} \\
     NYSED 
     & 0.05 & 0.16 & 0.26
     & 0.16 & 0.28 & 0.34
     & 0.22 & 0.42 & 0.55
     & \textbf{0.27} & \textbf{0.53} & \textbf{0.63} \\
     PILB 
     & 0.16 & 0.31 & 0.39 
     & 0.20 & 0.32 & 0.37
     & \textbf{0.46} & 0.75 & \textbf{0.86}
     & 0.42 & \textbf{0.76} & \textbf{0.86} \\
     SBO
     &0.03  & 0.05 & 0.08
     & 0.04  & 0.08  & 0.11 
     & \textbf{0.11} & \textbf{0.27}  & \textbf{0.36}
     & \textbf{0.11} & \textbf{0.27} & 0.35 \\
     \midrule
     Retail
     & 0.38 & 0.57 & 0.65
     & 0.41 & \textbf{0.66} & \textbf{0.74}
     & 0.35 & 0.53 & 0.60
     & \textbf{0.44} & \textbf{0.66} & \textbf{0.74} \\
     Finance
     & 0.52 & 0.72 & 0.79
     & 0.48 & 0.72 & 0.80
     & 0.58 & 0.77 & \textbf{0.82}
     & \textbf{0.62} & \textbf{0.79} & \textbf{0.82} \\
     Tech
     & 0.40 & 0.54 & 0.60 
     & 0.32 & 0.49 & 0.56
     & \textbf{0.42} & \textbf{0.61} & \textbf{0.66}
     & \textbf{0.42} & 0.57 & 0.61 \\
     Tech-H
     & 0.29 & 0.58 & 0.60 
     & 0.29 & 0.53 & 0.59
     & \textbf{0.56} & \textbf{0.73} & \textbf{0.76}
     & 0.53 & 0.72 & \textbf{0.76}\\
     \bottomrule

    \end{tabular}
    \caption{TACO versus baseline methods on retrieval performance.}
    \label{tab:main concat}
    \vspace{-5mm}

\end{table*}

\subsection{Sensitivity Analysis}
We conduct a sensitivity analysis by varying the embedding method, embedding model, LLM model, and TACO hyperparameters. Due to the number of experiments and space limitations, we report results only on the first six public datasets.

\paragraph{Embedding Method}
We only embed the column descriptions and compare TACO over the two baselines. 
All other models and parameters remain unchanged.
Table~\ref{tab:change concat} shows that TACO consistently outperforms other methods.

\begin{table}[t]
    \centering
    \small
    \begin{tabular}{c|cc|cc|cc}
        \toprule
        dataset & \multicolumn{2}{c|}{SSCD} & \multicolumn{2}{c|}{S2-only}  & \multicolumn{2}{c}{TACO} \\
        & H@1 & H@5
        & H@1 & H@5 
        & H@1 & H@5  \\
        \midrule
        ASIS
        & 0.13 & 0.19
        & 0.27 & 0.50
        & \textbf{0.33} & \textbf{0.56} \\
        ATBI
        & 0.08 & 0.16 
        & 0.38 & 0.55
        & \textbf{0.40} & \textbf{0.73} \\
        CWO
        & 0.08 & 0.20
        & \textbf{0.28} & \textbf{0.70}
        & \textbf{0.28} & 0.61 \\
        KIS
        & 0.15 & 0.24
        & 0.37 & 0.63
        & \textbf{0.54} & \textbf{0.73} \\
        NPFM
        & 0.08 & 0.16
        & 0.27 & 0.47
        & \textbf{0.38} & \textbf{0.64} \\
        NTSB
        & 0.06 & 0.10
        & \textbf{0.22} & 0.39
        & 0.20 & \textbf{0.41} 
        \\
        \bottomrule
    \end{tabular}
    \caption{Retrieval performance when we embed only the descriptions.}

    \label{tab:change concat}
        \vspace{-5mm}

\end{table}

\paragraph{Embedding Models}
We change the embedding model from all-MiniLM-L12-v2 to all-MiniLM-L6-v2,
and keep the other setups the same.
Table~\ref{tab:change embedding} shows that TACO outperforms the other baselines under a different embedding model.
\begin{table}[t]
    \centering
    \small
    \begin{tabular}{c|cc|cc|cc}
        \toprule
        dataset & \multicolumn{2}{c|}{SSCD} & \multicolumn{2}{c|}{S2-only}  & \multicolumn{2}{c}{TACO} \\
        & H@1 & H@5
        & H@1 & H@5 
        & H@1 & H@5  \\
        \midrule
        ASIS
        & 0.14 & 0.24 
        & 0.27 & 0.51
        & \textbf{0.31} & \textbf{0.56} \\
        ATBI
        & 0.12 & 0.24 
        & 0.34 & 0.56 
        & \textbf{0.41} & \textbf{0.71} \\
        CWO
        & 0.07 & 0.28
        & 0.28 & 0.69
        & \textbf{0.34} & \textbf{0.70} \\
        KIS 
        & 0.17 & 0.27
        & 0.36 & 0.66
        & \textbf{0.47} & \textbf{0.75} \\
        NPFM 
        & 0.11 & 0.23
        & 0.25 & 0.44
        & \textbf{0.41} & \textbf{0.65} \\
        NTSB
        & 0.07 & 0.11
        & \textbf{0.23} & \textbf{0.42}
        & 0.18 & 0.38 
        \\
        \bottomrule
    \end{tabular}
    \caption{Retieval performance when we use all-MiniLM-L6-v2 as the embedding model.}
    \label{tab:change embedding}
\end{table}

\paragraph{LLM Model}
We use another LLM model gpt-4.1-mini, and compare the three methods.
Table~\ref{tab:change llm} shows the results of the three methods using a different LLM model and
TACO consistently outperforms the other baselines.
\begin{table}[t]
    \centering
    \small
    \begin{tabular}{c|cc|cc|cc}
        \toprule
        dataset & \multicolumn{2}{c|}{SSCD} & \multicolumn{2}{c|}{S2-only}  & \multicolumn{2}{c}{TACO} \\
        & H@1 & H@5
        & H@1 & H@5 
        & H@1 & H@5  \\
        \midrule
        ASIS
        & 0.16 & 0.41
        & 0.27 & 0.52
        & \textbf{0.30} & \textbf{0.60} \\
        ATBI
        & 0.22 & 0.47
        & 0.31 & 0.58
        & \textbf{0.43} & \textbf{0.73} \\
        CWO
        & 0.28 & 0.58
        & 0.24 & 0.62
        & \textbf{0.44} & \textbf{0.73} \\
        KIS
        & 0.30 & 0.64
        & 0.44 & 0.72
        & \textbf{0.47} & \textbf{0.78} \\
        NPFM
        & 0.23 & 0.46
        & 0.34 & 0.59
        & \textbf{0.44} & \textbf{0.72} \\
        NTSB 
        & 0.16 & 0.35
        & \textbf{0.22} & 0.39
        & 0.20 & \textbf{0.42} 
        \\
        \bottomrule  
    \end{tabular}
    \caption{Retrieval performance using a different LLM.}
    \label{tab:change llm}
\end{table}

\begin{figure*}[t]
  \centering
  \subfloat[]{\includegraphics[width=0.32\textwidth]{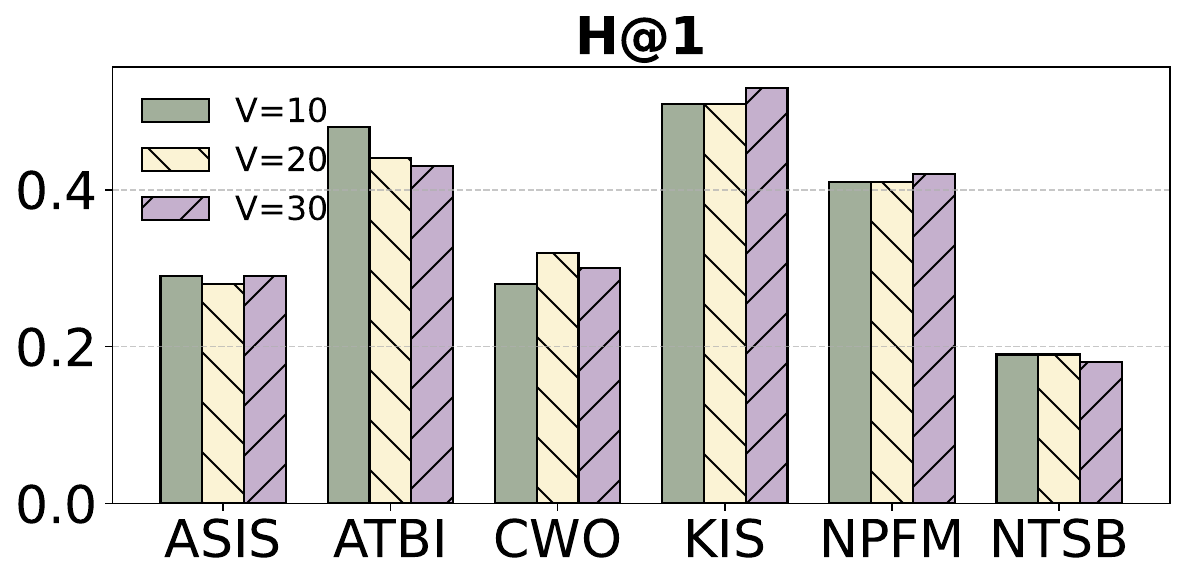}}
  \hfill
  \subfloat[]{\includegraphics[width=0.32\textwidth]{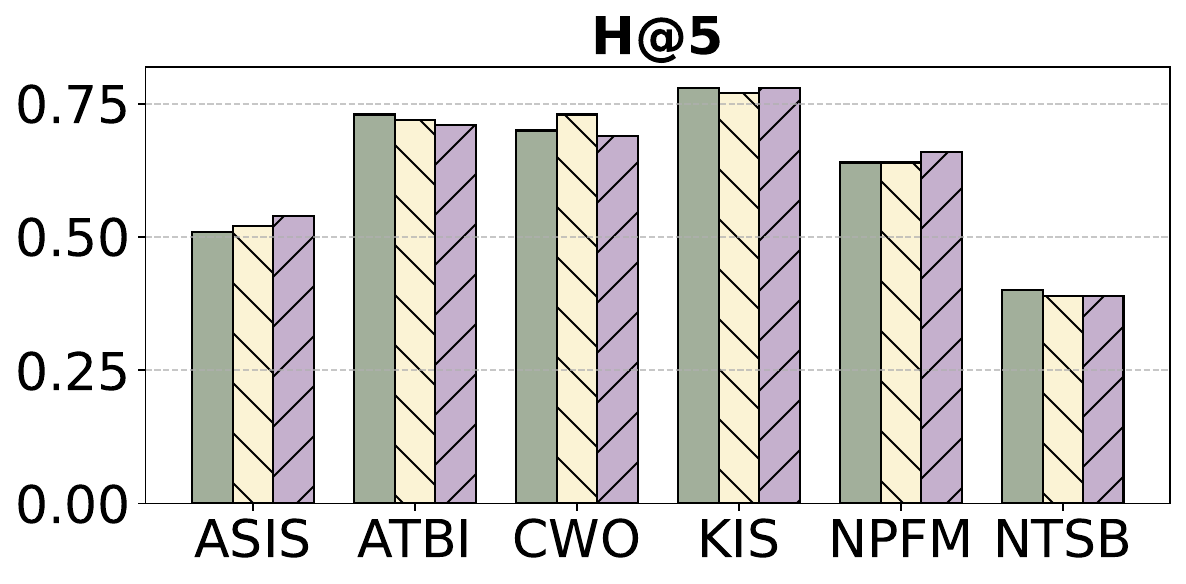}}
  \hfill
  \subfloat[]{\includegraphics[width=0.32\textwidth]
  {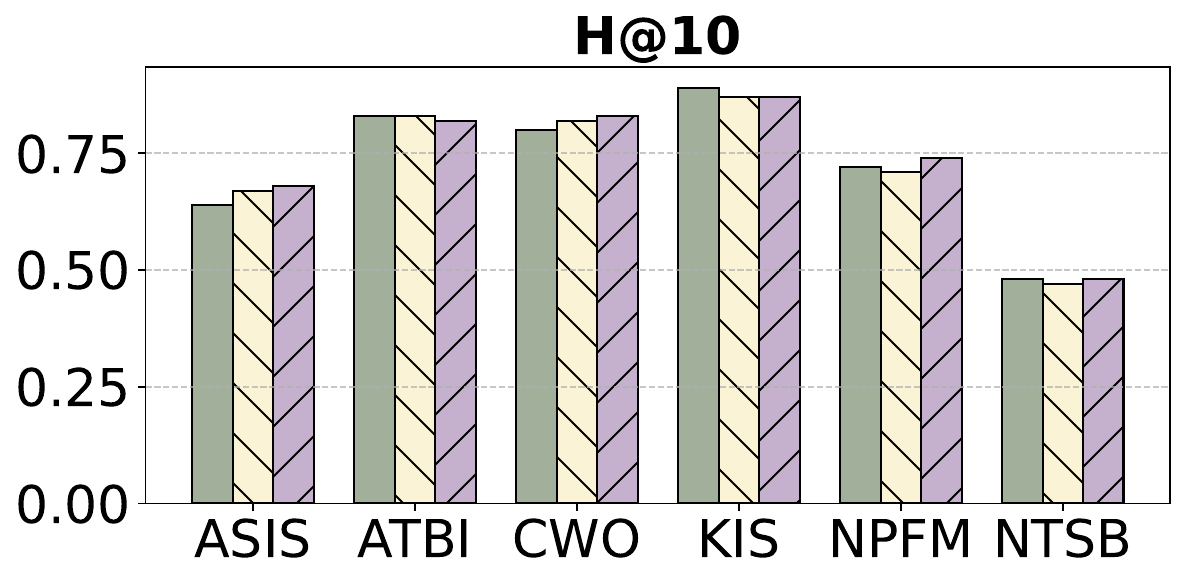}}
  \vspace{-5pt}
\caption{Retrieval performance as we vary 
$V$, the number of competing columns used for description revision.}
\label{fig:changeV}
\vspace{-15pt}
\end{figure*}

\paragraph{Hyperparameter $V$} 
In the description revision section, we use the simulated vector search results from the
competing top $V$ columns and ask LLM to revise the descriptions.
Currently, we use $V=30$.
In this section, we vary $V$ to be different numbers and report the results.
Figure~\ref{fig:changeV} shows that TACO is robust to different values of $V$.

\subsection{Ablation studies}
To demonstrate the effectiveness of TACO’s components, we perform an ablation study by removing each component in turn and comparing it with the full pipeline.
Table~\ref{tab:ablation} shows the results obtained by removing the expansion module and the revision module. The results show that both modules contribute meaningfully to TACO’s performance.

\begin{table}[t]
    \centering
    \small
        \vspace{-4mm}

    \begin{tabular}{c|cc|cc|cc}
        \toprule
        dataset & \multicolumn{2}{c|}{No-Exp} & \multicolumn{2}{c|}{No-Rev}  & \multicolumn{2}{c}{TACO} \\
        & H@1 & H@5
        & H@1 & H@5 
        & H@1 & H@5  \\
        \midrule
        ASIS
        & \textbf{0.33} & \textbf{0.59}
        & 0.29 & 0.49
        & \textbf{0.33} & 0.56 \\
        ATBI
        & 0.33 & 0.60
        & \textbf{0.40} & 0.66 
        & \textbf{0.40} & \textbf{0.73} \\
        CWO
        & 0.30 & \textbf{0.70}
        & \textbf{0.31} & 0.69 
        & 0.28 & 0.61 \\
        KIS 
        & 0.35 & 0.69
        & 0.48 & 0.72
        & \textbf{0.54} & \textbf{0.73}\\
        NPFM 
        & 0.27 & 0.48
        & 0.32 & 0.63
        & \textbf{0.38} & \textbf{0.64} \\
        NTSB
        & \textbf{0.21} & 0.39
        & \textbf{0.21} & 0.39
        & 0.20 & \textbf{0.41}
        \\
        \bottomrule
    \end{tabular}
    \caption{Ablation study on TACO. No-Exp removes the expansion step, and No-Rev removes the revision step.}
    \label{tab:ablation}
\end{table}

\subsection{Additional Human Input}\label{label:human}
\begin{figure*}[t]
  \centering
  \subfloat[]{\includegraphics[width=0.32\textwidth]{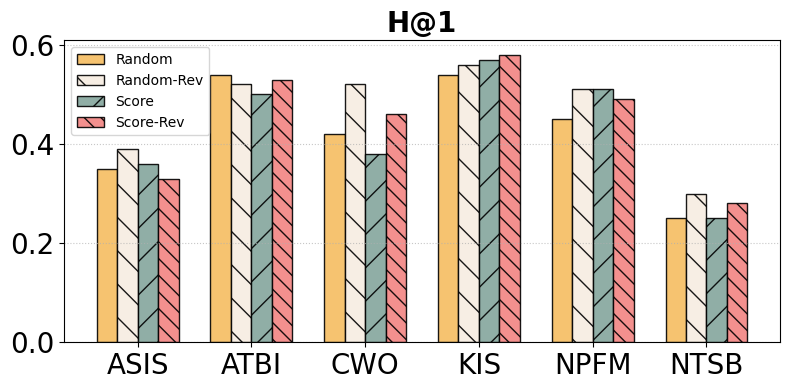}}
  \hfill
  \subfloat[]{\includegraphics[width=0.32\textwidth]{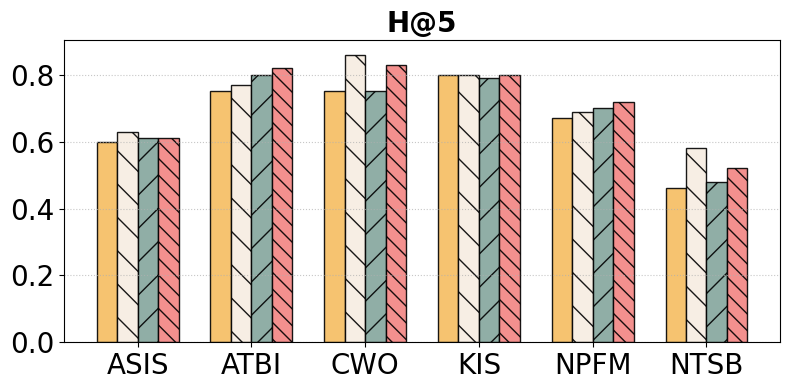}}
  \hfill
  \subfloat[]{\includegraphics[width=0.32\textwidth]
  {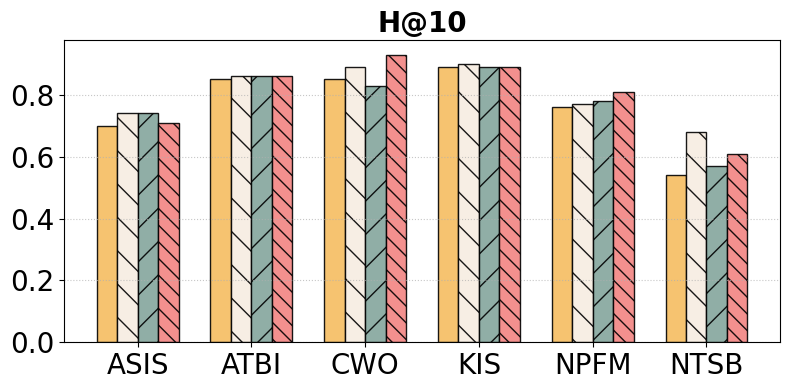}}
  \vspace{-5pt}
\caption{Retrieval performance using different human-in-the-loop strategies.}
\label{fig:human input}
\vspace{-15pt}
\end{figure*}

We evaluate the impact of incorporating additional human input for column and table name expansions. We compare four strategies:
(i) \textbf{Random}: We randomly select 10\% of columns and provide their correct column and table name expansions.
(ii) \textbf{Score}: We evaluates expansion correctness and select the 10\% of column names with the highest uncertainty for human verification by an LLM.
(iii) \textbf{Random-Rev}: After randomly curating 10\% of columns, we provide the human feedback to the LLM to revise the remaining column expansions.
(iv) \textbf{Score-Rev}: After selecting column names based on uncertainty scores, we use the human-provided expansions to guide the LLM in revising the remaining column names.
Figure~\ref{fig:human input} shows that random sampling and score-based sampling have similar performance, where the revising step helps both methods.

\subsection{TACO for non-abbreviated column names}
We also consider the setting where column names contain no abbreviations. Since abbreviation expansion is unnecessary, we remove Step 1 and directly apply description generation followed by the revision module. Table~\ref{tab:regular} reports results comparing three methods under this setting, where both column and table names have no abbreviations. TACO outperforms the other baselines by up to 7\%.
 
\begin{table}[t]
    \centering
    \small
    \begin{tabular}{c|cc|cc|cc}
    \toprule
        dataset  & \multicolumn{2}{c|}{SSCD} & \multicolumn{2}{c|}{S2-only}  & \multicolumn{2}{c}{TACO} \\
        & H@1 & H@5
        & H@1 & H@5 
        & H@1 & H@5  \\
        \midrule
        ASIS
        & 0.62 & 0.88 
        & 0.68 & 0.90
        & \textbf{0.71} & \textbf{0.95} \\
        ATBI
        & 0.76 & 0.96
        & \textbf{0.89} & 0.98
        & 0.84 & \textbf{1.00} \\
        CWO
        & 0.79 & 0.97
        & 0.77 & \textbf{0.99}
        & \textbf{0.83} & \textbf{0.99} \\
        KIS 
        & 0.70 & 0.90
        & 0.79 & \textbf{0.97}
        & \textbf{0.82} & \textbf{0.97} \\
        NPFM 
        & 0.74 & 0.95
        & 0.85 & 0.97
        & \textbf{0.86} & \textbf{0.98} \\
        NTSB
        & 0.52 & 0.84
        & \textbf{0.60} & \textbf{0.88}
        & 0.59 & \textbf{0.88} \\
        \bottomrule
        
    \end{tabular}
    \caption{Comparison on non-abbreviated names.}
    \label{tab:regular}\vspace{-5mm}

\end{table}



\section{Related Work}

\paragraph{Column Description Generation.}
Prior work has investigated generating natural language descriptions for table schemas. 
\citet{wretblad2024synthetic} focus on column description generation using a single LLM prompt, 
evaluating performance on NL2SQL tasks and through human assessment.  
Several recent efforts extend this line to both column and table descriptions 
\cite{zhang2025autoddg, tabmeta, singh2025leveragingretrievalaugmentedgenerative, 
10.14778/3748191.3748222,
10.1145/3708359.3712083,
gao2025automaticdatabasedescriptiongeneration}.  
For example, \citet{zhang2025autoddg} generate table-level descriptions to support readability and search, 
with column descriptions serving as supporting context.  
\citet{tabmeta} propose generating both table and column descriptions 
and employ an LLM-as-a-judge framework to identify high-quality outputs.  
\citet{singh2025leveragingretrievalaugmentedgenerative} generate descriptions for a proprietary dataset, 
assuming the availability of curated business glossaries; their approach handles abbreviations 
by aligning against these glossaries.  
\citet{gao2025automaticdatabasedescriptiongeneration} further scale the task to generate database-level descriptions.  
These works demonstrate the utility of schema descriptions, but they generally treat column names at face value 
and rely on either glossaries or single-prompt generation without addressing abbreviation-heavy real-world settings.  

\paragraph{Abbreviation Expansion.}
NameGuess \cite{nameguess} has explicitly addressed abbreviated column names in tables, 
introducing a benchmark with synthetic abbreviations and showing that LLMs can outperform 
fine-tuned models in one-shot prompting.  
Columbo~\cite{columbo} improves upon NameGuess and introduces a more robust framework to solve column name expansion
and evaluate their method over several new benchmarks.
Follow-up work \cite{anonymous2025realistic} explores generating more realistic abbreviations, 
\citet{snails} analyze how varying levels of abbreviation affect downstream tasks like NL2SQL 
without proposing an expansion method.  
 
\paragraph{Table Understanding and Enrichment.}
\blue{Column name expansion is a special case of the broader challenge of table understanding \cite{reviewer_cite3, reviewer_cite4, nguyen2025improving, reviewer_cite6} and enrichment. 
This includes enhancing data understanding \cite{fang2024large}, 
facilitating discovery \cite{freire2025large, reviewer_cite1, reviewer_cite2}, 
and supporting reasoning tasks such as table-to-text generation 
\cite{zhao2023investigating, zhao2023qtsumm, tabgenie, tableformer, gong19table} 
and table question answering \cite{pal-etal-2023-multitabqa, xie-etal-2022-unifiedskg, herzig-etal-2020-tapas}.}
Another related area is semantic type inference, where systems classify columns into predefined categories 
such as \emph{zip code}, \emph{address}, or \emph{date} 
\cite{turl22, deuer2024ca, sherlock19, duduo21, sato20, autotag21, adatyper23}. 

Other related tasks include schema matching and ontology alignment, 
where LLMs are leveraged to align columns across heterogeneous datasets 
or to external knowledge graphs \cite{semtab, yang2025matching, vandemoortele2024scalable}.

\section{Conclusion}
We presented \textsc{TACO}, 
a task-aware framework for column description generation that combines abbreviation expansion, 
schema-informed description generation, and revision to produce accurate and distinctive descriptions. 
Experiments show consistent improvements over baselines across multiple datasets, benefiting semantic retrieval tasks.
Our work demonstrates the challenges posed by abbreviation-heavy schemas and the necessity of addressing them in practical systems. These contributions emphasize the importance of robust schema interpretation and open up future research directions, including the integration of cell-level information and domain-specific knowledge resources to further improve semantic understanding.

\section*{Limitations}
Our evaluation dataset is not fully comprehensive, 
as queries generated for a given column may also be relevant to other semantically related columns. 
For instance, given a column name salary in the employee table, 
an LLM-generated query such as ``what is the average salary for all employees?'' 
could also pertain to related columns like base or stock that describe components of employee compensation. 
Constructing a truly complete dataset would therefore require substantial manual effort to 
capture such cross-column relationships. 
Nevertheless, to our knowledge no public benchmark currently exists for this task, 
and we release our dataset as an initial resource for the community, 
with plans to extend it into a more comprehensive benchmark in future work.

\bibliography{columbo, termmatcher}

@inproceedings{anonymous2025realistic,
  title={Realistic Training Data Generation and Rule Enhanced Decoding in {LLM} for NameGuess},
  author={Anonymous},
  booktitle={Submitted to ACL Rolling Review - December 2024},
  year={2025},
  url={https://openreview.net/forum?id=p9HwzYjFLe},
  note={under review} }

@article{semtab,
  title={Matching table metadata with business glossaries using large language models},
  author={Lobo, Elita and Hassanzadeh, Oktie and Pham, Nhan and Mihindukulasooriya, Nandana and Subramanian, Dharmashankar and Samulowitz, Horst},
  journal={arXiv preprint arXiv:2309.11506},
  year={2023}
}

@inproceedings{nameguess,
    title = "{N}ame{G}uess: Column Name Expansion for Tabular Data",
    author = "Zhang, Jiani  and
      Shen, Zhengyuan  and
      Srinivasan, Balasubramaniam  and
      Wang, Shen  and
      Rangwala, Huzefa  and
      Karypis, George",
    booktitle = "Proceedings of the 2023 Conference on Empirical Methods in Natural Language Processing",
    month = dec,
    year = "2023",
    address = "Singapore",
    publisher = "Association for Computational Linguistics",
    url = "https://aclanthology.org/2023.emnlp-main.820/",
    doi = "10.18653/v1/2023.emnlp-main.820",
    pages = "13276--13290"
}

@article{snails,
author = {Luoma, Kyle and Kumar, Arun},
title = {SNAILS: Schema Naming Assessments for Improved LLM-Based SQL Inference},
year = {2025},
issue_date = {February 2025},
publisher = {Association for Computing Machinery},
address = {New York, NY, USA},
volume = {3},
number = {1},
url = {https://doi.org/10.1145/3709727},
doi = {10.1145/3709727},
journal = {Proc. ACM Manag. Data},
month = feb,
articleno = {77},
numpages = {26}
}

@article{birdbench,
  title={Can llm already serve as a database interface? a big bench for large-scale database grounded text-to-sqls},
  author={Li, Jinyang and Hui, Binyuan and Qu, Ge and Yang, Jiaxi and Li, Binhua and Li, Bowen and Wang, Bailin and Qin, Bowen and Geng, Ruiying and Huo, Nan and others},
  journal={Advances in Neural Information Processing Systems},
  volume={36},
  year={2024}
}

@article{turl22,
  author       = {Xiang Deng and
                  Huan Sun and
                  Alyssa Lees and
                  You Wu and
                  Cong Yu},
  title        = {{TURL:} Table Understanding through Representation Learning},
  journal      = {{SIGMOD} Rec.},
  volume       = {51},
  number       = {1},
  pages        = {33--40},
  year         = {2022},
  url          = {https://doi.org/10.1145/3542700.3542709},
  doi          = {10.1145/3542700.3542709},
  bibsource    = {dblp computer science bibliography, https://dblp.org}
}

@misc{columbo,
      title={Columbo: Expanding Abbreviated Column Names for Tabular Data Using Large Language Models}, 
      author={Ting Cai and Stephen Sheen and AnHai Doan},
      year={2025},
      eprint={2508.09403},
      archivePrefix={arXiv},
      primaryClass={cs.CL},
      url={https://arxiv.org/abs/2508.09403}, 
}

@misc{singh2025leveragingretrievalaugmentedgenerative,
      title={Leveraging Retrieval Augmented Generative LLMs For Automated Metadata Description Generation to Enhance Data Catalogs}, 
      author={Mayank Singh and Abhijeet Kumar and Sasidhar Donaparthi and Gayatri Karambelkar},
      year={2025},
      eprint={2503.09003},
      archivePrefix={arXiv},
      primaryClass={cs.IR},
      url={https://arxiv.org/abs/2503.09003}, 
}

@misc{gao2025automaticdatabasedescriptiongeneration,
      title={Automatic database description generation for Text-to-SQL}, 
      author={Yingqi Gao and Zhiling Luo},
      year={2025},
      eprint={2502.20657},
      archivePrefix={arXiv},
      primaryClass={cs.AI},
      url={https://arxiv.org/abs/2502.20657}, 
}

@article{zhang2025autoddg,
  title={AutoDDG: Automated Dataset Description Generation using Large Language Models},
  author={Zhang, Haoxiang and Liu, Yurong and Santos, A{\'e}cio and Freire, Juliana and others},
  journal={arXiv preprint arXiv:2502.01050},
  year={2025}
}

@article{wretblad2024synthetic,
  title={Synthetic SQL Column Descriptions and Their Impact on Text-to-SQL Performance},
  author={Wretblad, Niklas and Holmstr{\"o}m, Oskar and Larsson, Erik and Wiks{\"a}ter, Axel and S{\"o}derlund, Oscar and {\"O}hman, Hjalmar and Pont{\'e}n, Ture and Forsberg, Martin and S{\"o}rme, Martin and Heintz, Fredrik},
  journal={arXiv preprint arXiv:2408.04691},
  year={2024}
}

@inproceedings{tabmeta,
  title={TabMeta: Table Metadata Generation with {LLM}-Curated Dataset and {LLM}-Judges},
  author={Anonymous},
  booktitle={Submitted to ACL Rolling Review - June 2024},
  year={2024},
  url={https://openreview.net/forum?id=NXYVm3AjG2},
  note={under review}
}

@inproceedings{zhao2023investigating,
  title={Investigating table-to-text generation capabilities of large language models in real-world information seeking scenarios},
  author={Zhao, Yilun and Zhang, Haowei and Si, Shengyun and Nan, Linyong and Tang, Xiangru and Cohan, Arman},
  booktitle={Proceedings of the 2023 Conference on Empirical Methods in Natural Language Processing: Industry Track},
  pages={160--175},
  year={2023}
}

@inproceedings{zhao2023qtsumm,
  title={{QTS}umm: Query-Focused Summarization over Tabular Data},
  author={Yilun Zhao and Zhenting Qi and Linyong Nan and Boyu Mi and Yixin Liu and Weijin Zou and SIMENG HAN and RUIZHE CHEN and Xiangru Tang and Yumo Xu and Dragomir Radev and Arman Cohan},
  booktitle={The 2023 Conference on Empirical Methods in Natural Language Processing},
  year={2023},
  url={https://openreview.net/forum?id=ubXaboYnzN}
}

@inproceedings{tabgenie,
    title = "{T}ab{G}enie: A Toolkit for Table-to-Text Generation",
    author = "Kasner, Zden{\v{e}}k  and
      Garanina, Ekaterina  and
      Platek, Ondrej  and
      Dusek, Ondrej",
    booktitle = "Proceedings of the 61st Annual Meeting of the Association for Computational Linguistics (Volume 3: System Demonstrations)",
    month = jul,
    year = "2023",
    address = "Toronto, Canada",
    publisher = "Association for Computational Linguistics",
    url = "https://aclanthology.org/2023.acl-demo.42/",
    doi = "10.18653/v1/2023.acl-demo.42",
    pages = "444--455"
}

@inproceedings{tableformer,
    title = "{T}able{F}ormer: Robust Transformer Modeling for Table-Text Encoding",
    author = "Yang, Jingfeng  and
      Gupta, Aditya  and
      Upadhyay, Shyam  and
      He, Luheng  and
      Goel, Rahul  and
      Paul, Shachi",
    booktitle = "Proceedings of the 60th Annual Meeting of the Association for Computational Linguistics (Volume 1: Long Papers)",
    month = may,
    year = "2022",
    address = "Dublin, Ireland",
    publisher = "Association for Computational Linguistics",
    url = "https://aclanthology.org/2022.acl-long.40/",
    doi = "10.18653/v1/2022.acl-long.40",
    pages = "528--537"
}

@inproceedings{gong19table,
    title = "Table-to-Text Generation with Effective Hierarchical Encoder on Three Dimensions (Row, Column and Time)",
    author = "Gong, Heng  and
      Feng, Xiaocheng  and
      Qin, Bing  and
      Liu, Ting",
    booktitle = "Proceedings of the 2019 Conference on Empirical Methods in Natural Language Processing and the 9th International Joint Conference on Natural Language Processing (EMNLP-IJCNLP)",
    month = nov,
    year = "2019",
    address = "Hong Kong, China",
    publisher = "Association for Computational Linguistics",
    url = "https://aclanthology.org/D19-1310/",
    doi = "10.18653/v1/D19-1310",
    pages = "3143--3152"
}

@inproceedings{pal-etal-2023-multitabqa,
    title = "{M}ulti{T}ab{QA}: Generating Tabular Answers for Multi-Table Question Answering",
    author = "Pal, Vaishali  and
      Yates, Andrew  and
      Kanoulas, Evangelos  and
      de Rijke, Maarten",
    booktitle = "Proceedings of the 61st Annual Meeting of the Association for Computational Linguistics (Volume 1: Long Papers)",
    month = jul,
    year = "2023",
    address = "Toronto, Canada",
    publisher = "Association for Computational Linguistics",
    url = "https://aclanthology.org/2023.acl-long.348/",
    doi = "10.18653/v1/2023.acl-long.348",
    pages = "6322--6334"
}

@inproceedings{xie-etal-2022-unifiedskg,
    title = "{U}nified{SKG}: Unifying and Multi-Tasking Structured Knowledge Grounding with Text-to-Text Language Models",
    author = "Xie, Tianbao  and
      Wu, Chen Henry  and
      Shi, Peng  and
      Zhong, Ruiqi  and
      Scholak, Torsten  and
      Yasunaga, Michihiro  and
      Wu, Chien-Sheng  and
      Zhong, Ming  and
      Yin, Pengcheng  and
      Wang, Sida I.  and
      Zhong, Victor  and
      Wang, Bailin  and
      Li, Chengzu  and
      Boyle, Connor  and
      Ni, Ansong  and
      Yao, Ziyu  and
      Radev, Dragomir  and
      Xiong, Caiming  and
      Kong, Lingpeng  and
      Zhang, Rui  and
      Smith, Noah A.  and
      Zettlemoyer, Luke  and
      Yu, Tao",
    booktitle = "Proceedings of the 2022 Conference on Empirical Methods in Natural Language Processing",
    month = dec,
    year = "2022",
    address = "Abu Dhabi, United Arab Emirates",
    publisher = "Association for Computational Linguistics",
    url = "https://aclanthology.org/2022.emnlp-main.39/",
    doi = "10.18653/v1/2022.emnlp-main.39",
    pages = "602--631"
}

@inproceedings{herzig-etal-2020-tapas,
    title = "{T}a{P}as: Weakly Supervised Table Parsing via Pre-training",
    author = {Herzig, Jonathan  and
      Nowak, Pawel Krzysztof  and
      M{\"u}ller, Thomas  and
      Piccinno, Francesco  and
      Eisenschlos, Julian},
    booktitle = "Proceedings of the 58th Annual Meeting of the Association for Computational Linguistics",
    month = jul,
    year = "2020",
    address = "Online",
    publisher = "Association for Computational Linguistics",
    url = "https://aclanthology.org/2020.acl-main.398/",
    doi = "10.18653/v1/2020.acl-main.398",
    pages = "4320--4333"
}

@article{fang2024large,
  title={Large Language Models (LLMs) on Tabular Data: Prediction, Generation, and Understanding--A Survey},
  author={Fang, Xi and Xu, Weijie and Tan, Fiona Anting and Zhang, Jiani and Hu, Ziqing and Qi, Yanjun and Nickleach, Scott and Socolinsky, Diego and Sengamedu, Srinivasan and Faloutsos, Christos},
  journal={arXiv preprint arXiv:2402.17944},
  year={2024}
}

@inproceedings{yang2025matching,
  title={Matching Table Metadata to Knowledge Graphs: A Data Augmentation Perspective},
  author={Duo Yang and Ioannis Dasoulas and Anastasia Dimou},
  booktitle={ELLIS workshop on Representation Learning and Generative Models for Structured Data},
  year={2025},
  url={https://openreview.net/forum?id=nvT5b9rvzf}
}

@misc{vandemoortele2024scalable,
  title={Scalable Table-to-Knowledge Graph Matching from Metadata using LLMs},
  author={Vandemoortele, Nathan and Steenwinckel, Bram and Hoecke, SV and Ongenae, Femke},
  year={2024},
  publisher={SemTab}
}

@article{freire2025large,
  author       = {Juliana Freire and
                  Grace Fan and
                  Benjamin Feuer and
                  Christos Koutras and
                  Yurong Liu and
                  Eduardo Pe{\~{n}}a and
                  A{\'{e}}cio S. R. Santos and
                  Cl{\'{a}}udio T. Silva and
                  Eden Wu},
  title        = {Large Language Models for Data Discovery and Integration: Challenges
                  and Opportunities},
  journal      = {{IEEE} Data Eng. Bull.},
  volume       = {49},
  number       = {1},
  pages        = {3--31},
  year         = {2025},
  url          = {http://sites.computer.org/debull/A25mar/p3.pdf},
  bibsource    = {dblp computer science bibliography, https://dblp.org}
}

@article{privacy_regulation2016regulation,
  title={Regulation (EU) 2016/679 of the European Parliament and of the Council},
  author={Regulation, Protection},
  journal={Regulation (eu)},
  volume={679},
  number={2016},
  pages={10--13},
  year={2016}
}

@article{privacy_illman2019california,
  title={California consumer privacy act},
  author={Illman, Erin and Temple, Paul},
  journal={The Business Lawyer},
  volume={75},
  number={1},
  pages={1637--1646},
  year={2019},
  publisher={JSTOR}
}

@inproceedings{schema1_zhang2023schema,
  title={Schema matching using pre-trained language models},
  author={Zhang, Yunjia and Floratou, Avrilia and Cahoon, Joyce and Krishnan, Subru and M{\"u}ller, Andreas C and Banda, Dalitso and Psallidas, Fotis and Patel, Jignesh M},
  booktitle={2023 IEEE 39th International Conference on Data Engineering (ICDE)},
  pages={1558--1571},
  year={2023},
  organization={IEEE}
}

@inproceedings{schema2_lei2020re,
  title={Re-examining the Role of Schema Linking in Text-to-SQL},
  author={Lei, Wenqiang and Wang, Weixin and Ma, Zhixin and Gan, Tian and Lu, Wei and Kan, Min-Yen and Chua, Tat-Seng},
  booktitle={Proceedings of the 2020 Conference on Empirical Methods in Natural Language Processing (EMNLP)},
  pages={6943--6954},
  year={2020}
}

@article{schema3_maamari2024death,
  title={The death of schema linking? text-to-sql in the age of well-reasoned language models},
  author={Maamari, Karime and Abubaker, Fadhil and Jaroslawicz, Daniel and Mhedhbi, Amine},
  journal={arXiv preprint arXiv:2408.07702},
  year={2024}
}

@inproceedings{reimers-2019-sentence-bert,
  title = "Sentence-BERT: Sentence Embeddings using Siamese BERT-Networks",
  author = "Reimers, Nils and Gurevych, Iryna",
  booktitle = "Proceedings of the 2019 Conference on Empirical Methods in Natural Language Processing",
  month = "11",
  year = "2019",
  publisher = "Association for Computational Linguistics",
  url = "https://arxiv.org/abs/1908.10084",
}

@article{reviewer_cite1,
  title={CHORUS: foundation models for unified data discovery and exploration},
  author={Kayali, Moe and Lykov, Anton and Fountalis, Ilias and Vasiloglou, Nikolaos and Olteanu, Dan and Suciu, Dan},
  journal={arXiv preprint arXiv:2306.09610},
  year={2023}
}

@article{reviewer_cite2,
  title={Mind the data gap: Bridging llms to enterprise data integration},
  author={Kayali, Moe and Wenz, Fabian and Tatbul, Nesime and Demiralp, {\c{C}}a{\u{g}}atay},
  journal={arXiv preprint arXiv:2412.20331},
  year={2024}
}

@article{reviewer_cite3,
  title={Chain-of-table: Evolving tables in the reasoning chain for table understanding},
  author={Wang, Zilong and Zhang, Hao and Li, Chun-Liang and Eisenschlos, Julian Martin and Perot, Vincent and Wang, Zifeng and Miculicich, Lesly and Fujii, Yasuhisa and Shang, Jingbo and Lee, Chen-Yu and others},
  journal={arXiv preprint arXiv:2401.04398},
  year={2024}
}

@article{reviewer_cite4,
  title={Tabsqlify: Enhancing reasoning capabilities of llms through table decomposition},
  author={Nahid, Md Mahadi Hasan and Rafiei, Davood},
  journal={arXiv preprint arXiv:2404.10150},
  year={2024}
}

@article{nguyen2025improving,
  title={Improving table understanding with LLMs and entity-oriented search},
  author={Nguyen, Thi-Nhung and Ngo, Hoang and Phung, Dinh and Vu, Thuy-Trang and Nguyen, Dat Quoc},
  journal={arXiv preprint arXiv:2508.17028},
  year={2025}
}

@inproceedings{reviewer_cite6,
  title={Hitab: A hierarchical table dataset for question answering and natural language generation},
  author={Cheng, Zhoujun and Dong, Haoyu and Wang, Zhiruo and Jia, Ran and Guo, Jiaqi and Gao, Yan and Han, Shi and Lou, Jian-Guang and Zhang, Dongmei},
  booktitle={Proceedings of the 60th annual meeting of the association for computational linguistics (volume 1: long papers)},
  pages={1094--1110},
  year={2022}
}

@article{autotag21,
  author       = {Yeye He and
                  Jie Song and
                  Yue Wang and
                  Surajit Chaudhuri and
                  Vishal Anil and
                  Blake Lassiter and
                  Yaron Goland and
                  Gaurav Malhotra},
  title        = {Auto-Tag: Tagging-Data-By-Example in Data Lakes},
  year         = {2021},
  url          = {https://arxiv.org/abs/2112.06049},
  bibsource    = {dblp computer science bibliography, https://dblp.org}
}

@article{duduo21,
  author       = {Yoshihiko Suhara and
                  Jinfeng Li and
                  Yuliang Li and
                  Dan Zhang and
                  {\c{C}}agatay Demiralp and
                  Chen Chen and
                  Wang{-}Chiew Tan},
  title        = {Annotating Columns with Pre-trained Language Models},
  journal      = {CoRR},
  volume       = {abs/2104.01785},
  year         = {2021},
  url          = {https://arxiv.org/abs/2104.01785},
  eprinttype    = {arXiv},
  eprint       = {2104.01785},
  bibsource    = {dblp computer science bibliography, https://dblp.org}
}

@inproceedings{sherlock19,
  author       = {Madelon Hulsebos and
                  Kevin Zeng Hu and
                  Michiel A. Bakker and
                  Emanuel Zgraggen and
                  Arvind Satyanarayan and
                  Tim Kraska and
                  {\c{C}}agatay Demiralp and
                  C{\'{e}}sar A. Hidalgo},
  title        = {Sherlock: {A} Deep Learning Approach to Semantic Data Type Detection},
  booktitle    = {Proceedings of the 25th {ACM} {SIGKDD} International Conference on
                  Knowledge Discovery {\&} Data Mining, {KDD} 2019},
  pages        = {1500--1508},
  publisher    = {{ACM}},
  year         = {2019},
  url          = {https://doi.org/10.1145/3292500.3330993},
  doi          = {10.1145/3292500.3330993},
  bibsource    = {dblp computer science bibliography, https://dblp.org}
}

@article{sato20,
  author       = {Dan Zhang and
                  Yoshihiko Suhara and
                  Jinfeng Li and
                  Madelon Hulsebos and
                  {\c{C}}agatay Demiralp and
                  Wang{-}Chiew Tan},
  title        = {Sato: Contextual Semantic Type Detection in Tables},
  journal      = {Proc. {VLDB} Endow.},
  volume       = {13},
  number       = {11},
  pages        = {1835--1848},
  year         = {2020},
  url          = {http://www.vldb.org/pvldb/vol13/p1835-zhang.pdf},
  bibsource    = {dblp computer science bibliography, https://dblp.org}
}

@article{adatyper23,
  author       = {Madelon Hulsebos and
                  Paul Groth and
                  {\c{C}}agatay Demiralp},
  title        = {AdaTyper: Adaptive Semantic Column Type Detection},
  year         = {2023},
  url          = {https://doi.org/10.48550/arXiv.2311.13806},
  doi          = {10.48550/ARXIV.2311.13806},
  eprinttype    = {arXiv},
  eprint       = {2311.13806},
  bibsource    = {dblp computer science bibliography, https://dblp.org}
}

@article{10.14778/3748191.3748222,
author = {Zhang, Tianshu and Qian, Kun and Sahai, Siddhartha and Tian, Yuan and Garg, Shaddy and Sun, Huan and Li, Yunyao},
title = {Evoschema: Towards Text-to-SQL Robustness against Schema Evolution},
year = {2025},
issue_date = {June 2025},
publisher = {VLDB Endowment},
volume = {18},
number = {10},
issn = {2150-8097},
url = {https://doi.org/10.14778/3748191.3748222},
doi = {10.14778/3748191.3748222},
journal = {Proc. VLDB Endow.},
month = jun,
pages = {3655–3668},
numpages = {14}
}

@inproceedings{10.1145/3708359.3712083,
author = {Tian, Yuan and Lee, Daniel and Wu, Fei and Mai, Tung and Qian, Kun and Sahai, Siddhartha and Zhang, Tianyi and Li, Yunyao},
title = {Text-to-SQL Domain Adaptation via Human-LLM Collaborative Data Annotation},
year = {2025},
isbn = {9798400713064},
publisher = {Association for Computing Machinery},
address = {New York, NY, USA},
url = {https://doi.org/10.1145/3708359.3712083},
doi = {10.1145/3708359.3712083},
booktitle = {Proceedings of the 30th International Conference on Intelligent User Interfaces},
pages = {1398–1425},
numpages = {28},
location = {
},
series = {IUI '25}
}

@inproceedings{10.1145/3654777.3676368,
author = {Tian, Yuan and Kummerfeld, Jonathan K. and Li, Toby Jia-Jun and Zhang, Tianyi},
title = {SQLucid: Grounding Natural Language Database Queries with Interactive Explanations},
year = {2024},
isbn = {9798400706288},
publisher = {Association for Computing Machinery},
address = {New York, NY, USA},
url = {https://doi.org/10.1145/3654777.3676368},
doi = {10.1145/3654777.3676368},
booktitle = {Proceedings of the 37th Annual ACM Symposium on User Interface Software and Technology},
articleno = {12},
numpages = {20},
location = {Pittsburgh, PA, USA},
series = {UIST '24}
}

@inproceedings{tian-etal-2023-interactive,
    title = "Interactive Text-to-{SQL} Generation via Editable Step-by-Step Explanations",
    author = "Tian, Yuan  and
      Zhang, Zheng  and
      Ning, Zheng  and
      Li, Toby Jia-Jun  and
      Kummerfeld, Jonathan K.  and
      Zhang, Tianyi",
    editor = "Bouamor, Houda  and
      Pino, Juan  and
      Bali, Kalika",
    booktitle = "Proceedings of the 2023 Conference on Empirical Methods in Natural Language Processing",
    month = dec,
    year = "2023",
    address = "Singapore",
    publisher = "Association for Computational Linguistics",
    url = "https://aclanthology.org/2023.emnlp-main.1004/",
    doi = "10.18653/v1/2023.emnlp-main.1004",
    pages = "16149--16166"
}

@article{deuer2024ca,
  author       = {Benjamin Feuer and
                  Yurong Liu and
                  Chinmay Hegde and
                  Juliana Freire},
  title        = {ArcheType: {A} Novel Framework for Open-Source Column Type Annotation
                  using Large Language Models},
  journal      = {Proc. {VLDB} Endow.},
  volume       = {17},
  number       = {9},
  pages        = {2279--2292},
  year         = {2024},
  url          = {https://www.vldb.org/pvldb/vol17/p2279-freire.pdf},
  bibsource    = {dblp computer science bibliography, https://dblp.org}
}

\newpage
\section*{Appendix}
\subsection{SNAILS SBOD dataset}
Table~\ref{tab:sbod dataset} shows the details of the sub-datasets in the SBOD dataset from SNALS~\cite{snails}.
\setlength{\tabcolsep}{3pt}
\begin{table}[t]
    \centering
    \small
    \begin{tabular}{cccc}
        \toprule
    dataset & \# tables & \# columns & \# queries\\
    \midrule
     Banking & 40 & 1758 & 1758  \\
      Business Partners & 40 & 1456 & 1456 \\
      Finance & 61 & 1994 & 1994 \\
      General & 71 & 1053 & 1053 \\
      Human Resources & 28 & 455 & 455 \\
      Inventory and Production & 65 & 1960 & 1960 \\
      Reports & 40 & 744 & 744 \\
      Sales Opportunities & 20 & 290 & 290 \\
      Service & 40 & 902 & 902
      \\
      \bottomrule
    \end{tabular}
    \caption{9 datasets in SBOD}
    \label{tab:sbod dataset}
\end{table}

\subsection{Incorporating Table Description}
\label{apx: incorporate table description}
To demonstrate that TACO is an extensible framework,
we add a new module to generate table description before the step 2 (column description generation),
and use the table description as the input for the column description as well.
Table~\ref{tab:table desc} shows the performance when we add the table description module to TACO.
\begin{table}[t]
    \centering
    \small
    \begin{tabular}{c|ccc|ccc}
        \toprule
        dataset & \multicolumn{3}{c|}{TACO} &  \multicolumn{3}{c}{TACO+Table Desc} \\
        & H@1 & H@5 & H@10
        & H@1 & H@5 & H@10 \\
        \midrule
        ASIS
        & 0.29 & 0.54 & \textbf{0.68}
        & \textbf{0.30} & \textbf{0.56} & \textbf{0.68} \\
        ATBI
        & 0.43 & 0.71 & \textbf{0.82}
        & \textbf{0.47} & \textbf{0.76} & \textbf{0.82} \\
        CWO
        & 0.30 & 0.69 & \textbf{0.83}
        & \textbf{0.34} & \textbf{0.72} & 0.77 \\
        KIS
        & \textbf{0.53} & \textbf{0.78} & \textbf{0.87}
        & 0.50 & 0.76 & \textbf{0.87} \\
        NPFM
        & \textbf{0.42} & \textbf{0.66} & 0.74
        & 0.41 & \textbf{0.66} & \textbf{0.75}
        \\
        NTSB
        & 0.18 & \textbf{0.39} & \textbf{0.48}
        & \textbf{0.19} & \textbf{0.39} & \textbf{0.48}
        \\
        \bottomrule
    \end{tabular}
    \caption{add another module for table description before generating the column description}
    \label{tab:table desc}
\end{table}

\subsection{Prompt of Table Name Expansion}
\label{sec:table name expansion prompt}
Figure~\ref{fig:table name expansion} shows the prompt for table name expansion.
\begin{figure*}
    \centering
\begin{tcolorbox}[
  colback=gray!5,
  colframe=gray!50,
  width=\linewidth,
  boxrule=0.5pt,
  arc=2mm,
  left=2mm,
  right=2mm,
  top=1mm,
  bottom=1mm
]
\begin{lstlisting}[style=prompt]
Given the following information about a table in the {domain} area:
Table Name: {table_name}
Column Names: {column_names}
Column Expansion (generated by LLM, may be wrong): {column_expansion}

Can you expand the table name into full form English words or phrases? Follow the following rules when you expand:
1. do not add any extra information or explanation in your final expansion.
2. Do not mutate the numbers appear in the attribute names.
3. Keep the original orders of the attribute names.
4. If a token is already in full form, its expansion should be itself and do not paraphrase it.

First anaylze the context of the table and explain your expansion of the table name.
At last, output your answer in a JSON format where the key is the original table name and the value is the expanded table name.
\end{lstlisting}
\end{tcolorbox}
    \caption{Prompt for Table Name Expansion}
    \label{fig:table name expansion}
\end{figure*}

\subsection{Prompt of Column Description Generation}
\label{sec:column description prompt}
Figure~\ref{fig:column desc generation} shows the prompt for column description generation. The \{context\} includes the randomly sampled column names from the same table.
\begin{figure*}
    \centering
\begin{tcolorbox}[
  colback=gray!5,
  colframe=gray!50,
  width=\linewidth,
  boxrule=0.5pt,
  arc=2mm,
  left=2mm,
  right=2mm,
  top=1mm,
  bottom=1mm
]
\begin{lstlisting}[style=prompt]
You are a helpful assistant. Your task is to generate descriptions for attributes in the {domain} vertical. These descriptions aim to help downstream search engines to identify related attributes towards user queries.

The attributes are all from the same table, the table has the following information:
- table name: {table_name}
- column names: {context}
For each attribute, you will be provided with the attribute name.

Your job is to generate a description for the attribute that is concise, informative, non-ambiguous, relevant to the domain and help the downstream search engines to understand the attribute.
1. Moreover, you should also ensure that you do not directly paraphrase the original description nor simply explain the attribute.
2. You should try to describe the attribute name to make it more easily to be searched.
3. You should describe the attribute name in a whole and not just describe a single part.
4. Do not hallucinate or make up any ifnormation that is not in the attribute name, altdisplay, original_desc or expansion.

Apart from the description, you should also try to generate the following information helpful for downstream search engines:
1. a list of keywords that are relevant to the attribute, separated by commas.
2. a list of synonyms that are relevant to the attribute, separated by commas.
3. a list of potential search queries that are relevant to the attribute, separated by commas.

When you generate the keywords, synonyms and search queries, make sure
1. they are relevant to the attribute and the domain.
2. don't generate vague or generic keywords, synonyms or search queries.
3. make it very specific to the attribute and the domain.

The attribute information is as follows: attribute_name: {column_name}

Can you generate the description, keywords, synonyms and search queries for each of the attribute?
Please return the result in a JSON format.
Also can you do a self revision:
1. first explain your understanding of the attribute and the task.
2. then generate the description, keywords, synonyms and search queries.
3. finally, pinpoint the drawbacks of your generated description, keywords, synonyms and search queries and revise them if necessary.
4. return the final result in a JSON format where the key is the column name (make sure it is in the exact format of the input column names), the value is a dictionary with keys being descriptions, keywords, synonyms, and search queries.
\end{lstlisting}
\end{tcolorbox}
    \caption{Prompt for column description generation}
    \label{fig:column desc generation}
\end{figure*}

\subsection{Prompt of Column Description Revision}
\label{apx:column revision}
Figure~\ref{fig:query generation} shows the prompt for the simulated query generation.
Figure~\ref{fig:desc rev} shows the prompt to revise the generated descriptions based on simulated downstream tasks feedback.

\begin{figure*}
    \centering
    \begin{tcolorbox}[
  colback=gray!5,
  colframe=gray!50,
  width=\linewidth,
  boxrule=0.5pt,
  arc=2mm,
  left=2mm,
  right=2mm,
  top=1mm,
  bottom=1mm
]
\begin{lstlisting}[style=prompt]
Given the examples below (each attribute is in table_name.column_name format), can you generate queries for my input attribute? You can generate 3 queries for each input attribute:

attribute: tblFieldDataTurtleMeasurements.age
query: How many five year old turtles were measured?

attribute: tblFieldDataTurtleMeasurements.LocationlD
query: How many turtles were measured at each location where turtles were measured?

attribute: tblFieldDataTurtleMeasurements.Weight
query: what is the average weight of all turtles?

attribute: tblFieldDataSnakeDataCollection.SVL
query: what is the highest snake snout-to-vent length recorded?

attribute: tblFieldDataGreenCardObservations.Species_Code
query: how many distinct species were documented on a reptile survey green card?

attribute: tblFieldDataTurtleTrapSurveys.Trap_Type
query: How many records are logged for each turtle trap type?

attribute: tblFieldDataTurtleMeasurements.gravid
query: how many turtle measurements were of turtles carrying eggs or young?

Input attributes: {insert}
Output in a JSON format where the keys are the attribute name and the values are the list of generated queries. The format should be:
### final result
{
  "placeholder_attribute_name": [
    "place_holder_query_ 1 ",
    "place_holder_query_2",
    "place_holder_query_3"
  ]
}
\end{lstlisting}
\end{tcolorbox}
    \caption{Prompt for column query generation}
    \label{fig:query generation}
\end{figure*}

\begin{figure*}
    \centering
\begin{tcolorbox}[
  colback=gray!5,
  colframe=gray!50,
  width=\linewidth,
  boxrule=0.5pt,
  arc=2mm,
  left=2mm,
  right=2mm,
  top=1mm,
  bottom=1mm
]
\begin{lstlisting}[style=prompt]
Given the column name: {answer} (which is the correct column name for the following queries), when embed only on the LLM generated descripions of this column name and perform vector search, the correct column name is not returned as the first rank in the result.

The actual description of the column is: {actual_description}
Given the following results from the vector search:
Query: {query}
Rank of the column name: {rank}
Top 30 vector search result: {result}

1. Can you explain in detail why the correct column is not rank the first in the returned results compared to other column's descriptions?
2. Can you focus on how to revise the description (together with the synonyms, keywords, search_queries) for the column '{answer}' so that its cosine similairty between the query increases and the vector search result is better?

Provide the 3 most relevant synonyms, keywords and user queries. Return the revised enrichment as the same JSON format.
\end{lstlisting}
\end{tcolorbox}
    \caption{Prompt for description revision using results from simulated queries}
    \label{fig:desc rev}
\end{figure*}

\subsection{Prompt of Human Input}
\label{apx:sec:prompt human input}
Figure~\ref{fig:llm judge} shows the prompt for using LLM-as-a-judge to evaluate the column expansions from step 1.
Figure~\ref{fig:human revise exp} shows the prompt to re-generate the column expansions based on human input.

\begin{figure*}
    \centering
\begin{tcolorbox}[
  colback=gray!5,
  colframe=gray!50,
  width=\linewidth,
  boxrule=0.5pt,
  arc=2mm,
  left=2mm,
  right=2mm,
  top=1mm,
  bottom=1mm
]
\begin{lstlisting}[style=prompt]
Given the following information about a table in the {domain} area:
Table Name: {table_name}
Table Expansion {generated by LLM, may be wrong): {table_expansion}
Column Name and Column Expansion {expansions are generated by LLM, may be wrong):
{column_name}: {column_expansion}

Task:
For the table name and each column name:
1. Assign a confidence score from O to 5 indicating how likely you think the provided expansion is correct, based on your knowledge and the information given:
0: Very likely incorrect; almost certainly wrong.
1: Likely incorrect; probably wrong, very little matches.
2: Possibly incorrect; some clues, but mostly doubtful.
3: Uncertain; about equally likely to be correct or incorrect.
4: Likely correct; mostly matches your understanding, but with minor doubts.
5: Almost certainly correct; matches standard usage, very high confidence.
2. Briefly explain your reasoning for the assigned score.

Output the result in a JSON format, where the keys are the table name and the column names (keep them in the original format) and the value is a dictionary with the following keys
1. expansion: the provided expansion,
2. score: the ambiguity score,
3. reason: a short reason for the score

\end{lstlisting}
\end{tcolorbox}
    \caption{Prompt for LLM-as-a-judge for the column expansions}
    \label{fig:llm judge}
\end{figure*}

\begin{figure*}
    \centering
\begin{tcolorbox}[
  colback=gray!5,
  colframe=gray!50,
  width=\linewidth,
  boxrule=0.5pt,
  arc=2mm,
  left=2mm,
  right=2mm,
  top=1mm,
  bottom=1mm
]
\begin{lstlisting}[style=prompt]
Given the following information about a table in the {domain} area:
Table Name: {table_name}
Table Expansion (generated by LLM, may be wrong): {table_expansion}
Column Name and Column Expansion (expansions are generated by LLM, may be wrong):
{column_name}: {column_expansion}

Given the following user feedback, in the format of
[
  original_table_name.original_column_name, expanded_table_name.expanded_column_name
]:
1. user expansion from the same table: {user_feedback_from_the_same_table}
2. user expansion from other tables: {user_feedback_from_the_other_table}
Please revise the table expansion and column expansions based on the user feedback.
Return the revised table expansion and column expansions in the JSON format, where the keys are the original table name and column names (do not add table name in front of the column names in the keys) and the values are the revised expansions.
\end{lstlisting}
\end{tcolorbox}
    \caption{Prompt for re-generating column expansions based on retrieved human input expansions}
    \label{fig:human revise exp}
\end{figure*}

\end{document}